\documentclass[10pt,twocolumn,letterpaper]{article}

\usepackage{cvpr}
\usepackage{times}
\usepackage{epsfig}
\usepackage{graphicx}
\usepackage{amsmath}
\usepackage{color}
\usepackage{amssymb}
\usepackage{booktabs}
\usepackage{epigraph}
\usepackage{xcolor}            % color

\usepackage[pagebackref=true,breaklinks=true,letterpaper=true,colorlinks,bookmarks=false]{hyperref}

\newcommand{\reffig}[1]{Figure~\ref{fig:#1}}

\newcommand{\refsec}[1]{Section~\ref{sec:#1}}

\newcommand{\lblsec}[1]{\label{sec:#1}}

\definecolor{Plum}{RGB}{142, 69, 133}
\definecolor{Cyan}{RGB}{0, 255, 255}
\definecolor{Red3}{HTML}{a40000}
\definecolor{Green3}{HTML}{4e9a06}

\cvprfinalcopy % *** Uncomment this line for the final submission

\def\cvprPaperID{1621} % *** Enter the CVPR Paper ID here

\ifcvprfinal\fi
\begin{document}

%%%%%%%%% TITLE
\title{
Learning Dense Correspondence via 3D-guided Cycle Consistency
}

\author{Tinghui Zhou\\
UC Berkeley\\
\and
Philipp Kr\"ahenb\"uhl\\
UC Berkeley\\
\and
Mathieu Aubry\\
ENPC ParisTech\\
\and
Qixing Huang\\
TTI-Chicago\\
\and
Alexei A. Efros\\
UC Berkeley\\
}

\maketitle

\begin{abstract}
Discriminative deep learning approaches have shown impressive results for problems where human-labeled ground truth is plentiful, but what about tasks where labels are difficult or impossible to obtain?  This paper tackles one such problem: establishing dense visual correspondence across different object instances. For this task, although we do not know what the ground-truth is, we know it should be \textbf{consistent} across instances of that category. We exploit this consistency as a supervisory signal to train a convolutional neural network to predict cross-instance correspondences between pairs of images depicting objects of the same category.  For each pair of training images we find an appropriate 3D CAD model and render two synthetic views to link in with the pair, establishing a correspondence flow 4-cycle. We use ground-truth synthetic-to-synthetic correspondences, provided by the rendering engine, to train a ConvNet to predict synthetic-to-real, real-to-real and real-to-synthetic correspondences that are cycle-consistent with the ground-truth. At test time, no CAD models are required. We demonstrate that our end-to-end trained ConvNet supervised by cycle-consistency outperforms state-of-the-art pairwise matching methods in correspondence-related tasks.
\end{abstract}

\section{Introduction}

\setlength{\epigraphrule}{0pt}
\epigraph{\hspace{1.2in}{\em Consistency is all I ask!}}{{\sc Tom Stoppard}}

In the past couple of years, deep learning has swept though computer vision like wildfire.
One needs only to buy a GPU, arm oneself with enough training data, and turn the crank to see head-spinning improvements on most computer vision benchmarks.
So it is all the more curious to consider tasks
for which deep learning has {\em not} made much inroad, typically due to the lack of easily obtainable training data. 
One such task is {\em dense visual correspondence} -- the problem of estimating a pixel-wise correspondence field between images depicting visually similar objects or scenes. Not only is this a key ingredient for optical flow and stereo matching, but many other computer vision tasks, including recognition, segmentation, depth estimation, etc. could be posed as finding correspondences in a large visual database followed by label transfer.

In cases where the images depict the same physical object/scene across varying viewpoints, such as in stereo matching, there is exciting new work that aims to use the commonality of the scene structure as supervision to learn deep features for correspondence~\cite{agrawal2015learning,deepstereo,dinesh2015,han2015matchnet,vzbontar2015stereo}.   
But for computing correspondence {\em across different object/scene instances}, no learning method to date has managed to seriously challenge SIFT flow~\cite{liu2011sift}, the dominant approach for this task. 

\begin{figure}[t]
\centering
\includegraphics[width=\linewidth]{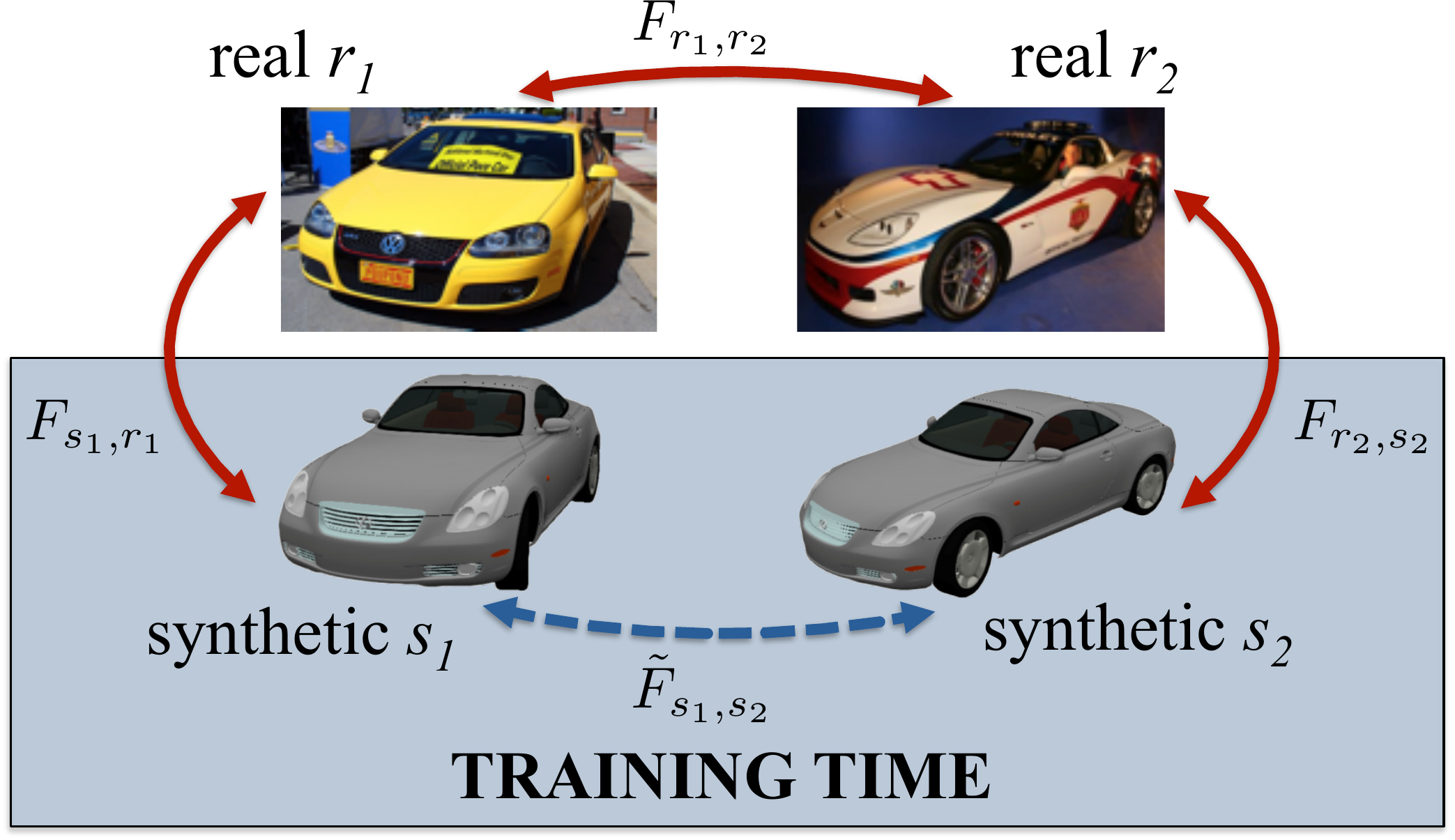}%scale=0.5
\caption{Estimating a dense correspondence flow field $F_{r_1,r_2}$ between two images $r_1$ and $r_2$  --- essentially, where do pixels of $r_1$ need to go to bring them into correspondence with $r_2$ --- is very difficult. There is a large viewpoint change, and the physical differences between the cars are substantial.  We propose to {\em learn} to do this task by training a ConvNet using the concept of cycle consistency in lieu of ground truth.  At training time, we find an appropriate 3D CAD model to establish a correspondence 4-cycle, and train the ConvNet to minimize the discrepancy between $\tilde{F}_{s_1\!, s_2}$ and $F_{s_1\!, r_1}\!\circ\!F_{r_1\!, r_2}\!\circ\!F_{r_2\!, s_2}$, where $\tilde{F}_{s_1\!, s_2}$ is known by construction.  At test time, no CAD models are used.}
\label{fig:quartet}
\end{figure}

How can we get supervision for dense correspondence between images depicting different object instances, such as images $r_1$ and $r_2$ in Figure~\ref{fig:quartet}?
Our strategy in this paper is to learn the things we don't know by linking them up to the things we do know. In particular, at training time, we use a large dataset of 3D CAD models~\cite{shapenet} to find one that could link the two images, as shown in Figure~\ref{fig:quartet}. Here the dense correspondence between the two views of the same 3D model $s_1$ and $s_2$ can serve as our ground truth supervision (as we know precisely where each shape point goes when rendered in a different viewpoint), but the challenge is to use this information to train a network that can produce correspondence between two real images at test time.

A naive strategy is to train a network to estimate correspondence between the rendered views of the same 3D model, and then hope that the network could generalize to real images as well. Unfortunately, this does not work in practice (see Table~\ref{tab:pck}), likely due to 1) the large visual difference between synthetic and real images and 2) the lack of cross-instance ground truth correspondence for training. Instead, in this paper we utilize the concept of {\em cycle consistency} of correspondence flows~\cite{huang2013consistent,zhou2015flowweb,zhou2015multi} -- the notion that the composition of flow fields for any circular path through the image set should have a zero combined flow. Here, cycle consistency serves as a way to link the correspondence between real images and the rendered views into a single 4-cycle chain.   We can then train our correspondence network using cycle consistency as the supervisory signal. The idea is to take advantage of the known synthetic-to-synthetic correspondence as ground-truth anchors that allow cycle consistency to propagate the correct correspondence information from synthetic to real images, without diverging or falling into a trivial solution. Here we could interpret the cycle consistency as a kind of ``meta-supervision'' that operates not on the data directly, but rather on how the data should behave.  As we show later, such 3D-guided consistency supervision allows the network to learn cross-instance correspondence that potentially overcomes some of the major difficulties (e.g. significant viewpoint and appearance variations) of previous pairwise matching methods like SIFT flow~\cite{liu2011sift}. Our approach could also be thought of as an extension and a  reformulation of FlowWeb~\cite{zhou2015flowweb} as a learning problem, where the image collection is stored implicitly in the network representation.

The main contributions of this paper are: 1) We propose a general learning framework for tasks without direct labels through cycle consistency as an example of ``meta-supervision''; 2) We present the first end-to-end trained deep network for dense cross-instance correspondence; 3) We demonstrate that the widely available 3D CAD models can be used for learning correspondence between 2D images of different object instances.

\section{Related work}
\noindent{\bf Cross-instance pairwise correspondence}
~~The classic SIFT Flow approach~\cite{liu2011sift} proposes an energy minimization framework that computes dense correspondence between different scenes by matching SIFT features~\cite{sift} regularized by smoothness and small displacement priors. Deformable Spatial Pyramid (DSP) Matching~\cite{kim2013deformable}, a recent follow-up to SIFT Flow, greatly speeds up the inference while modestly improving the matching accuracy. Barnes~\etal~\cite{barnes2010generalized} extend the original PatchMatch~\cite{barnes2009patchmatch} algorithm to allow more general-purpose (including cross-instance) matching. Bristow~\etal~\cite{bristow2015dense} build an exemplar-LDA classifier around each pixel, and aggregate the matching responses over all classifiers with additional smoothness priors to obtain dense correspondence estimation. In these same proceedings, Ham~\etal~\cite{ham2016} take advantage of recent developments in object proposals, and utilize local and geometric consistency constraints among object proposals to establish dense semantic correspondence.
\\\\
\noindent{\bf Collection correspondence}~~Traditionally, correspondence has been defined in a pairwise manner, but recent works have tried to pose correspondence as the problem of joint image-set alignment. The classic like on work on Congealing~\cite{learned2006data,huang2007unsupervised} uses sequential optimization to gradually lower the entropy of the intensity distribution of the entire image set by continuously warping each image via a parametric transformation (e.g. affine). RASL~\cite{peng2012rasl}, Collection Flow~\cite{kemelmacher2012collection} and Mobahi ~\etal~\cite{mobahi2014compositional} first estimate a low-rank subspace of the image collection, and then perform joint alignment among images projected onto the subspace. FlowWeb~\cite{zhou2015flowweb} builds a fully-connected graph for the image collection with images as nodes and pairwise flow fields as edges, and establishes globally-consistent dense correspondences by maximizing the cycle consistency among all edges. While achieving state-of-the-art performance, FlowWeb is overly dependent on the initialization quality, and scales poorly with the size of the image collection. Similar to a recent work on joint 3D shape alignment~\cite{huang2013consistent}, Zhou~\etal~\cite{zhou2015multi} tackle the problem by jointly optimizing feature matching and cycle consistency, but formulate it as a low-rank matrix recovery which they solve with a fast alternating minimization method. Virtual View Networks~\cite{carreira2014virtual} leverage annotated keypoints to infer dense correspondence between images connected in a viewpoint graph, and use this graph to align a query image to all the reference images in order to perform single-view 3D reconstruction. Cho~\etal~\cite{cho2015unsupervised} use correspondence consistency among selective search windows in a diverse image collection to perform unsupervised object discovery.
\\\\
\noindent{\bf Deep learning for correspondence}~~Recently, several works have applied convolutional neural networks to learn same-instance dense correspondence. FlowNet~\cite{fischer2015flownet} learns an optical flow CNN with a synthetic Flying Chairs dataset that generalizes well to existing benchmark datasets, yet still falls a bit short of state-of-the-art optical flow methods like DeepFlow~\cite{weinzaepfel2013deepflow} and EpicFlow~\cite{revaud2015epicflow}. Several recent works have also used supervision from reconstructed 3D scene and stereo pairs~\cite{han2015matchnet,vzbontar2015stereo,agrawal2015learning}. However all these approaches are inherently limited to matching images of the same physical object/scene. Long~\etal~\cite{long2014convnets} use deep features learned from large-scale object classification tasks to perform intra-class image alignment, but found it to perform similarly to SIFT flow.
\\\\
{\bf Image-shape correspondence}~~Our work is partially motivated by recent progress in image-shape alignment that allows establishing correspondence between images through intermediate 3D shapes. Aubry~\etal~\cite{Aubry:2014:SCE} learns discriminative patches for matching 2D images to their corresponding 3D CAD models, while Peng~\etal~\cite{peng2015learning} utilizes CAD models to train object detectors with few shots of labeled real images. In cases where depth data is available, deep learning methods have recently been applied to 3D object recognition and alignment between CAD models and RGB-D  images~\cite{guptaCVPR15a,song16,wu20153d}. Other works~\cite{DBLP:journals/tog/HuangWK15,shmlg_imageDepth_sig14} leverage image and shape collections for joint pose estimation and refining image-shape alignment, which are further applied to single-view object reconstruction and depth estimation. Although our approach requires 3D CAD models for constructing the training set, the image-shape alignment is jointly learned with the image-image alignment, and no CAD models are required at test time.

\section{Approach}
\begin{figure*}[t]
\centering
\includegraphics[width=\linewidth]{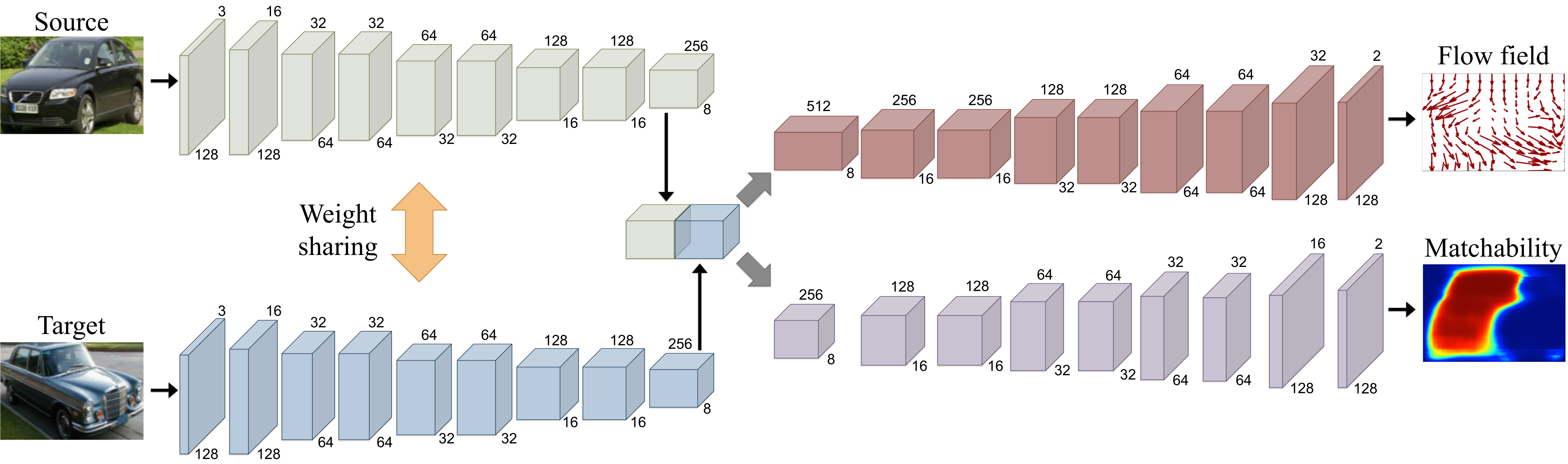}
\caption{Overview of our network architecture, which consists of three major components: 1) \textbf{feature encoder} on both input images, 2) \textbf{flow decoder} predicting the dense flow field from the source to the target image and 3) \textbf{matchability decoder} that outputs a probability map indicating whether each pixel in the source image has a correspondence in the target. See~\refsec{network} for more details.}
\label{fig:net}
\end{figure*}

Our goal is to predict a dense flow (or correspondence) field $F_{a,b} : \mathbb R^2 \to \mathbb R^2$ between pairs of images $a$ and $b$. The flow field $F_{a,b}(p) = (p_x - q_x, p_y - q_y)$ computes the relative offset from each point $p$ in image $a$ to a corresponding point $q$ in image $b$. Given that pairwise correspondence might not always be well-defined
(e.g. a side-view car and a frontal-view car do not have many visible parts in common), we additionally compute a matchability map $M_{a,b} : \mathbb R^2 \to [0,1]$ predicting if a correspondence exists $M_{a,b}(p)=1$ or not $M_{a,b}(p)=0$.

We learn both the flow field and the matchability prediction with a convolutional neural network. Both functions are differentiable with respect to the network parameters, which could be directly learned if we had dense annotations for $F_{a,b}$ and $M_{a,b}$ on a large set of real image pairs. However, in practice it is infeasible to obtain those annotations at scale as they are either too time-consuming or ambiguous to annotate. 

We instead choose a different route, and learn both functions by placing the supervision on the desired properties of the ground-truth, i.e.
while we do not know what the ground-truth is, we know how it should behave. In this paper, we use {\em cycle consistency} with 3D CAD models as the desired property that will be our supervisory signal. Specifically, for each pair of real training images $r_1$ and $r_2$, we find a 3D CAD model of the same category, and render two synthetic views $s_1$ and $s_2$ in similar viewpoint as $r_1$ and $r_2$, respectively (see \refsec{quartets} for more details). For each training quartet $<s_1,s_2,r_1,r_2>$ we learn to predict flows from $s_1$ to $r_1$ ($F_{s_1, r_1}$) to $r_2$ ($F_{r_1, r_2}$) to $s_2$ ($F_{r_2, s_2}$) that are cycle-consistent with the ground-truth flow from $s_1$ to $s_2$ ($\tilde{F}_{s_1, s_2}$) provided by the rendering engine (similarly for the matchability prediction).
By constructing consistency supervision through 3D CAD models, we aim to learn 2D image correspondences that potentially captures the 3D semantic appearance of the query objects.  Furthermore, making $\tilde{F}_{s_1, s_2}$ be ground-truth by construction prevents the cycle-consistency optimization from producing trivial solutions, such as identity flows.

Sections \ref{sec:objective} and \ref{sec:matchability} formally define our training objective for learning correspondence $F$ and matchability $M$, respectively. Section~\ref{sec:lerp} demonstrates how to obtain continuous approximation of discrete maps that allows end-to-end training. Section~\ref{sec:network} describes our network architecture.

\subsection{Learning dense correspondence}
\lblsec{objective}
Given a set of training quartets $\{<s_1, s_2, r_1, r_2>\}$, we train the CNN to minimize the following objective:
\begin{equation}
\sum_{<\!s_1\!,s_2\!,r_1\!,r_2\!>}\!\!\!\mathcal{L}_{flow}\left( \tilde{F}_{s_1\!, s_2},~F_{s_1\!, r_1}\!\circ\!F_{r_1\!, r_2}\!\circ\!F_{r_2\!, s_2} \right),\label{eq:obj}
\end{equation}
where $\tilde{F}_{s_1, s_2}$ refers to the ground-truth flow between two synthetic views, $F_{s_1,r_1}$, $F_{r_1, r_2}$ and $F_{r_2, s_2}$ are predictions made by the CNN along the transitive path. The transitive flow composition $\bar{F}_{a,c} = F_{a,b} \circ F_{b,c}$ is defined as 
\begin{equation}
\bar{F}_{a,c}(p) = F_{a,b}(p) + F_{b,c}(p+F_{a,b}(p))~,
\label{eq:comp}
\end{equation}
which is differentiable as long as $F_{a,b}$ and $F_{b,c}$ are differentiable. $\mathcal{L}_{flow}(\tilde{F}_{s_1, s_2}, \bar{F}_{s_1, s_2})$ denotes the truncated Euclidean loss defined as
\begin{align*}
\mathcal{L}_{flow} & (\tilde{F}_{s_1, s_2}, \bar{F}_{s_1, s_2})=\\
&\sum_{p | \tilde{M}_{s_1, s_2}(p)=1}\min(\|\tilde{F}_{s_1,s_2}(p) - \bar{F}_{s_1,s_2}(p)\|^2,T^2)~,
\end{align*}
where $\tilde{M}_{s_1, s_2}(p)$ is the ground-truth matchability map provided by the rendering engine ($\tilde{M}_{s_1, s_2}(p) = 0$ when $p$ is either a background pixel or not visible in $s_2$), and  $T=15$ (pixels) for all our experiments. In practice, we found the truncated loss to be more robust to spurious outliers for training, especially during the early stage when the network output tends to be highly noisy.

\subsection{Learning dense matchability}
\lblsec{matchability}
Our training objective for matchability prediction also utilizes the cycle consistency signal:
\begin{equation}
    \sum_{<\!s_1\!,s_2\!,r_1\!,r_2\!>}\!\!\!\mathcal{L}_{mat}\left( \tilde{M}_{s_1\!, s_2},~ M_{s_1\!, r_1}\!\circ\!M_{r_1\!, r_2}\!\circ\!M_{r_2\!, s_2} \right),\label{eq:mobj}
\end{equation}
where $\tilde{M}_{s_1,s_2}$ refers to the ground-truth matchability map between the two synthetic views, $M_{s_1,r_1}$, $M_{r_1, r_2}$ and $M_{r_2, s_2}$ are CNN predictions along the transitive path, and $\mathcal{L}_{mat}$ denotes per-pixel cross-entropy loss. The matchability map composition is defined as
\begin{equation}
    \bar{M}_{a,c}(p) = M_{a,b}(p) M_{b,c}(p+F_{a,b}(p))~,
    \label{eq:mapcomp}
\end{equation}
where the composition depends on both the matchability as well as the flow field.

Due to the multiplicative nature in matchability composition (as opposed to additive in flow composition), we found that training with objective~\ref{eq:mobj} directly results in the network exploiting the clean background in synthetic images, which helps predict a perfect segmentation of the synthetic object in $M_{s_1, r_1}$. Once $M_{s_1, r_1}$ predicts zero values for background points, the network has no incentive to correctly predict the matchability for background points in $M_{r_1,r_2}$, as the multiplicative composition has zero values regardless of the transitive predictions along $M_{r_1, r_2}$ and $M_{r_2, s_2}$.
To address this, we fix $M_{s_1, r_1}=\mathbf{1}$ and $M_{r_2, s_2}=\mathbf{1}$, and only train the CNN to infer $M_{r_1,r_2}$.
This assumes that every pixel in $s_1 (s_2)$ is matchable in $r_1 (r_2)$, and allows the matchability learning to happen between real images. Note that this is still different from directly using $\tilde{M}_{s_1, s_2}$ as supervision for $M_{r_1,r_2}$ as the matchability composition depends on the predicted flow field along the transitive path.

The matchability objective~\ref{eq:mobj} is jointly optimized with the flow objective~\ref{eq:obj} during training, and our final objective can be written as $\sum_{<\!s_1\!,s_2\!,r_1\!,r_2\!>}\mathcal{L}_{flow} + \lambda \mathcal{L}_{mat}$ with $\lambda = 100$.

\subsection{Continuous approximation of discrete maps}
\lblsec{lerp}
An implicit assumption made in our derivation of the transitive composition (Eq.~\ref{eq:comp} and~\ref{eq:mapcomp}) is that $F$ and $M$ are differentiable functions over continuous input, while images inherently consist of discrete pixel grids. To allow end-to-end training with stochastic gradient descent (SGD), we obtain continuous approximation of the full flow field and the matchability map with bilinear interpolation over the CNN predictions on discrete pixel locations. Specifically, for each discrete pixel location $\hat p \in \{1,\ldots,W\} \times \{1,\ldots,H\}$, the network predicts a flow vector $F_{a,b}(\hat p)$ as well as a matchability score $M_{a,b}(\hat p)$, and the approximation over all continuous points $p \in [1, W] \times [1, H]$ is obtained by:

\begin{align*}
 F_{a,b}(p) &= \sum_{\hat{p} \in \mathcal{N}_p} (1-|p_x-\hat{p}_x|)(1-|p_y-\hat{p}_y|) F_{a,b}(\hat{p}) \\
M_{a,b}(p) & = \sum_{\hat{p} \in \mathcal{N}_p} (1-|p_x-\hat{p}_x|)(1-|p_y-\hat{p}_y|) M_{a,b}(\hat{p})~,
\end{align*}
where $\mathcal{N}_p$ denotes the four-neighbor pixels (top-left, top-right, bottom-left, bottom-right) of point $p$, or just $p$ if it is one of the discrete pixels. This is equivalent to the differentiable image sampling with a bilinear kernel proposed in~\cite{stn}.

\subsection{Network architecture}
\lblsec{network}
Our network architecture (see \reffig{net}) follows the encoder-decoder design principle with three major components: 1) \textbf{feature encoder} of $8$ convolution layers that extracts relevant features from both input images with shared network weights; 2) \textbf{flow decoder} of $9$ fractionally-strided/up-sampling convolution (uconv) layers that assembles features from both input images, and outputs a dense flow field; 3) \textbf{matchability decoder} of $9$ uconv layers that assembles features from both input images, and outputs a probability map indicating whether each pixel in the source image has a correspondence in the target. 

All conv/uconv layers are followed by rectified linear units (ReLUs) except for the last uconv layer of either decoder, and the filter size is fixed to $3\times 3$ throughout the whole network. No pooling layer is used, and the stride is $2$ when increasing/decreasing the spatial dimension of the feature maps. The output of the matchability decoder is further passed to a sigmoid layer for normalization.

During training, we apply the same network to three different input pairs along the cycle ($s_1 \rightarrow r_1, r_1, \rightarrow r_2, $ and $r_2 \rightarrow s_2$), and composite the output to optimize the consistency objectives~\ref{eq:obj} and~\ref{eq:mobj}.

\section{Experimental Evaluation}
In this section, we describe the details of our network training procedure, and evaluate the performance of our network on correspondence and matchability tasks. 

\subsection{Training set construction}
\lblsec{quartets}
The 3D CAD models we used for constructing training quartets come from the ShapeNet database~\cite{shapenet}, while the real images are from the PASCAL3D+ dataset~\cite{xiang2014beyond}. For each object instance (cropped from the bounding box and rescaled to $128 \times 128$) in the train split of PASCAL3D+, we render all 3D models under the same camera viewpoint (provided by PASCAL3D+), and only use $K = 20$ nearest models as matches to the object instance based on the HOG~\cite{hog} Euclidean distance. We then construct training quartets each consisting of two real images ($r_1$ and $r_2$) matched to the same 3D model and their corresponding rendered views ($s_1$ and $s_2$). On average, the number of valid training quartets for each category is about $80,000$.

\subsection{Network training}
We train the network in a category-agnostic manner (i.e. a single network for all categories). We first initialize the network (feature encoder + flow decoder pathway) to mimic SIFT flow by randomly sampling image pairs from the training quartets and training the network to minimize the Euclidean loss between the network prediction and the SIFT flow output on the sampled pair\footnote{We also experimented with other initialization strategies (e.g. predicting ground-truth flows between synthetic images), and found that initializing with SIFT flow output works the best.}. Then we fine-tune the whole network end-to-end to minimize the consistency loss defined in Eq.~\ref{eq:obj} and~\ref{eq:mobj}. We use the ADAM solver~\cite{adam} with $\beta_1 = 0.9, \beta_2 = 0.999$, initial learning rate of $0.001$, step size of $50,000$, step multiplier of $0.5$ for $200,000$ iterations. We train with mini-batches of $40$ image pairs during initialization and $10$ quartets during fine-tuning.

\begin{figure*}[t]
\centering
\includegraphics[width=\linewidth]{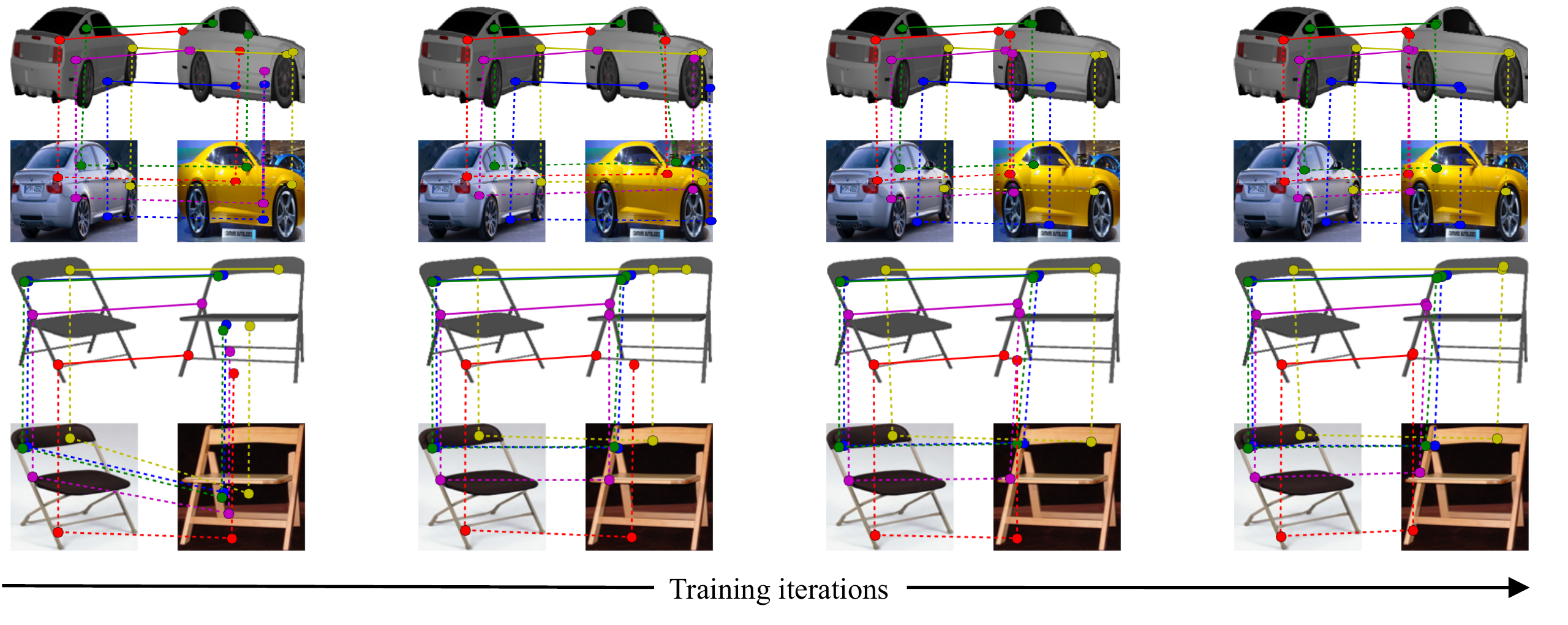}
\caption{Visualizing the effects of consistency training on the network output. The randomly sampled ground-truth correspondences between synthetic images are marked in solid lines, and the correspondence predictions along the cycle (synthetic to real, real to real and real to synthetic) made by our network are marked in dashed lines. One can see that the transitive composition of our network output becomes more and more consistent with the ground-truth as training progresses, while individual correspondences along each edge of the cycle also tend to become more semantically plausible.}
\label{fig:train}
\end{figure*}

We visualize the effect of our cycle-consistency training in Figure~\ref{fig:train}, where we sample some random points in the synthetic image $s_1$, and plot their predicted correspondences along the cycle $s_1 \rightarrow r_1 \rightarrow r_2 \rightarrow s_2$ to compare with the ground-truth in $s_2$. One can see that the transitive trajectories become more and more cycle-consistent with more iterations of training, while individual correspondences along each edge of the cycle also tend to become more semantically plausible.

\subsection{Feature visualization}
\begin{figure*}[t]
\centering
\includegraphics[width=\linewidth]{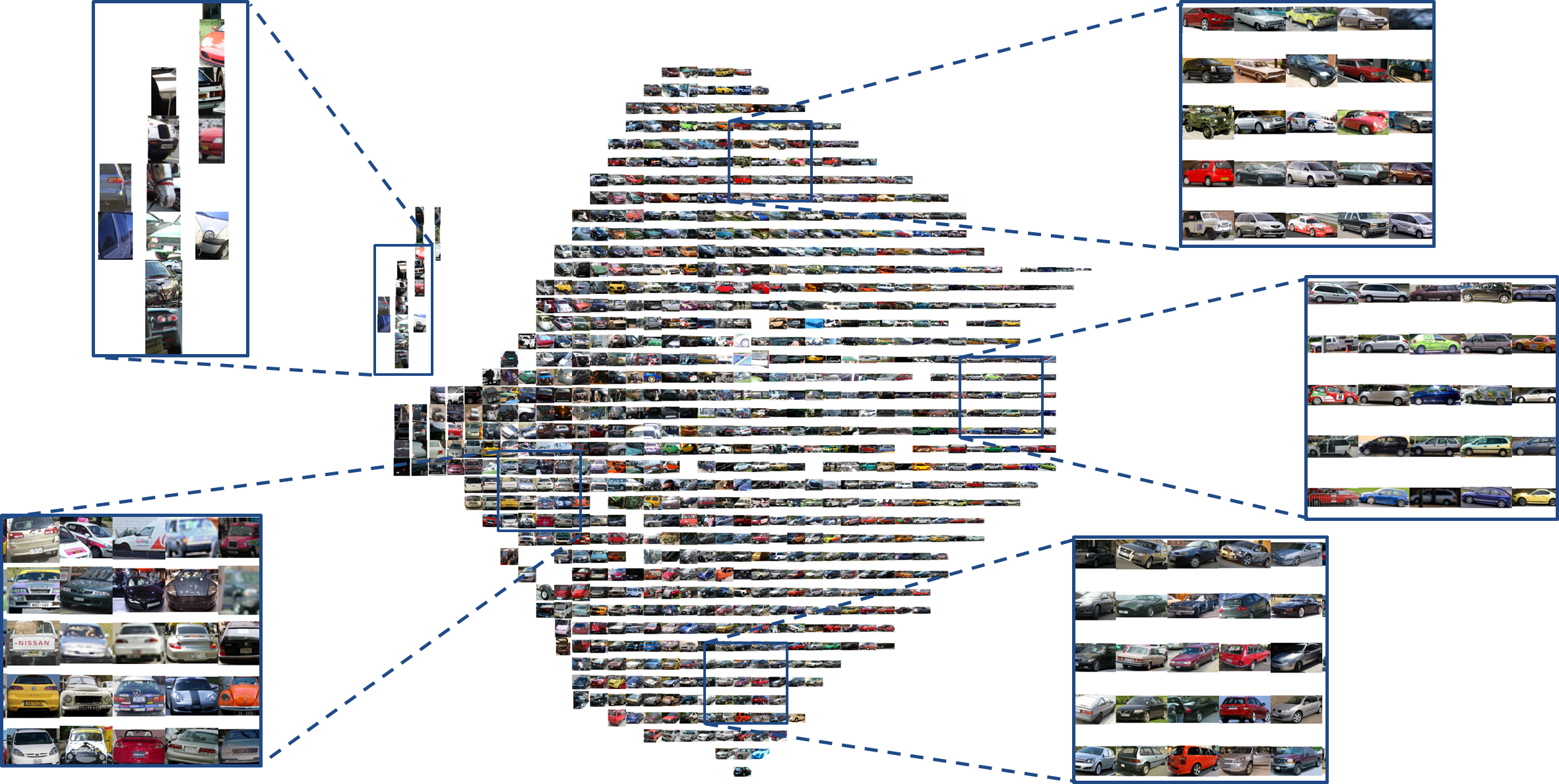}
\caption{Conv-9 feature embedding for cars visualized by t-SNE~\cite{tsne}. Interestingly, the overall layout seems to be mainly clustered based on the camera viewpoint, while the network is not explicitly trained to perform viewpoint estimation. This suggests that the network might implicitly learn that viewpoint is an important cue for the correspondence/matchability tasks through our consistency training.}
\label{fig:tsne}
\end{figure*}

We visualize the features learned by the network using the t-SNE algorithm~\cite{tsne}. Specifically, we extract conv-9 features (i.e. the output of the last encoder layer) from the entire set of car instances in the PASCAL3D+ dataset, and embed them in 2-D with the t-SNE algorithm. Figure~\ref{fig:tsne} visualizes the embedding. Interestingly, while our network is not explicitly trained to perform viewpoint estimation, the embedding layout appears to be viewpoint-sensitive, which implies that the network might implicitly learn that viewpoint is an important cue for correspondence/matchability tasks through our consistency training. 

\begin{figure*}[t]
\centering
\includegraphics[width=\linewidth]{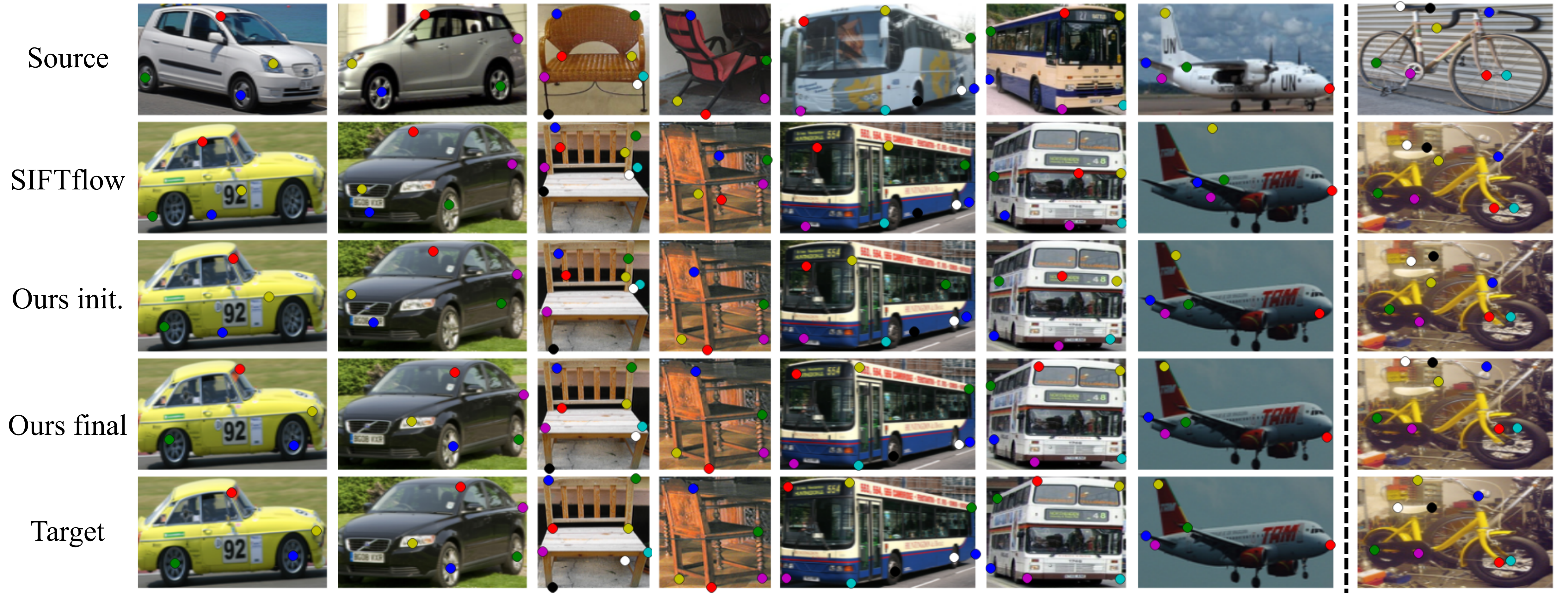}
\caption{Comparison of keypoint transfer performance for different methods on example test image pairs. Overall, our consistency-supervised network (second-to-last row) is able to produce more accurate keypoint transfer results than the baselines. The last column shows a case when SIFT flow performs better than ours.}
\label{fig:key}
\end{figure*}

\begin{table*}[t]
\centering
\scalebox{0.88}{
\begin{tabular}{ccccccccccccc|c}
\toprule
& aero & bike & boat & bottle & bus & car & chair & table & mbike & sofa & train & tv & {\bf mean} \tabularnewline
 \midrule
SIFT flow~\cite{liu2011sift} 
& $9.8$  & $\mathbf{23.3}$ & $8.9$  & $28.3$ & $28.6$ & $22.4$ & $10.8$ & $\mathbf{13.2}$ & $\mathbf{17.9}$ & $14.2$ & $14.4$ & $42.9$ & $19.6$ \tabularnewline
Long \etal~\cite{long2014convnets} 
& $10.4$ & $22.8$ & $7.6$  & $30.8$ & $28.4$ & $21.1$ & $10.2$ & $12.7$ & $13.5$ & $12.9$ & $12.6$ & $38.5$ & $18.5$ \tabularnewline
\midrule
$\text{CNN}_{I2S}$
& $9.1$  & $14.7$ & $5.2$  & $25.9$ & $25.4$ & $23.7$ & $11.9$ & $11.3$ & $13.4$ & $16.8$ & $11.3$ & $45.2$ & $17.8$ \tabularnewline
$\text{CNN}_{init}$ 
& $8.6$  & $20.3$ & $8.5$  & $29.4$ & $24.3$ & $20.1$ & $9.9$  & $11.6$ & $15.4$ & $11.6$ & $12.5$ & $40.2$ & $17.7$ \tabularnewline
$\text{CNN}_{init} + $ Synthetic ft.
& $10.2$ & $22.2$ & $8.7$  & $30.4$ & $24.5$ & $21.3$ & $10.2$ & $12.1$ & $15.7$ & $12.0$ & $12.8$ & $40.5$ & $18.4$ \tabularnewline
{\bf $\text{CNN}_{init} + $ Consistency ft.} 
& $\mathbf{11.3}$ & $22.3$ & $\mathbf{10.1}$ & $\mathbf{40.3}$ & $\mathbf{40.3}$ & $\mathbf{33.3}$ & $\mathbf{15.0}$ & $\mathbf{13.2}$ & 17.2 & $\mathbf{17.4}$ & $\mathbf{16.7}$ & $\mathbf{51.1}$ & $\mathbf{24.0}$ \tabularnewline
\bottomrule
\end{tabular}
}
\vspace{0.07in}
\caption{Keypoint transfer accuracy measured in PCK ($\alpha = 0.1$) on the PASCAL3D+ categories. Overall, our final network (last row) outperforms all baselines (except on ``bicycle'' and ``motorbike''). Notice the performance gap between our initialization ($\text{CNN}_{init}$) and the final network, which highlights the improvement made by cycle-consistency training.}
\label{tab:pck}
\end{table*}

\subsection{Keypoint transfer}
We evaluate the quality of our correspondence output using the keypoint transfer task on the $12$ categories from PASCAL3D+~\cite{xiang2014beyond}. For each category, we exhaustively sample all image pairs from the val split (not seen during training), and determine if a keypoint in the source image is transferred correctly by measuring the Euclidean distance between our correspondence prediction and the annotated ground-truth (if exists) in the target image. A correct transfer means the prediction falls within $\alpha \cdot \max(H,W)$ pixels of the ground-truth with $H$ and $W$ being the image height and width, respectively (both are $128$ pixels in our case). We compute the percentage of correct keypoint transfer (PCK) over all image pairs as the metric, and provide performance comparison for the following methods in Table~\ref{tab:pck}:
\begin{itemize}
\item SIFT flow~\cite{liu2011sift} -- A classic method for dense correspondence using SIFT feature descriptors and hand-designed smoothness and large-displacement priors. We also ran preliminary evaluation on a more recent follow-up based on deformable spatial pyramids~\cite{kim2013deformable}, and found it to perform similarly to SIFT flow.
\item Long \etal~\cite{long2014convnets} -- Similar MRF energy minimization framework as SIFT flow but with deep features learned from the ImageNet classification task.
\item CNN$_{I2S}$ -- Our network trained on real image pairs with correspondence inferred by compositing the output of an off-the-shelf image-to-shape alignment algorithm~\cite{DBLP:journals/tog/HuangWK15} and the ground-truth synthetic correspondence (i.e. obtaining direct supervision for $F_{r_1,r_2}$ through $F_{r_1,s_1} \circ \tilde{F}_{s_1, s_2} \circ F_{s_2,r_2},$ where $F_{r_1,s_1}$ and $F_{s_2,r_2}$ are inferred from~\cite{DBLP:journals/tog/HuangWK15}). 
\item CNN$_{init}$ -- Our network trained to mimic SIFT flow.
\item CNN$_{init} + $ Synthetic ft. -- fine-tuning on synthetic image pairs with ground-truth correspondence after initialization with SIFT flow.
\item CNN$_{init} + $ Consistency ft. -- fine-tuning with our objectives~\ref{eq:obj} and~\ref{eq:mobj} after initialization with SIFT flow.
\end{itemize}

Overall, our consistency-supervised network significantly outperforms all other methods (except on ``bicycle" and ``motorbike" where SIFT flow has a slight advantage). Notice the significant improvement over the initial network after consistency fine-tuning. The performance gap between the last two rows of Table~\ref{tab:pck} suggests that consistency supervision is much more effective in adapting to the real image domain than direct supervision from synthetic ground-truth.

Figure~\ref{fig:key} compares sample keypoint transfer results using different methods. In general, our final prediction tends to match the ground-truth much better than the other baselines, and could sometimes overcome substantial viewpoint and appearance variation where previous methods, like SIFT flow, are notoriously error-prone.

\begin{figure*}[t]
\centering
\includegraphics[width=\linewidth]{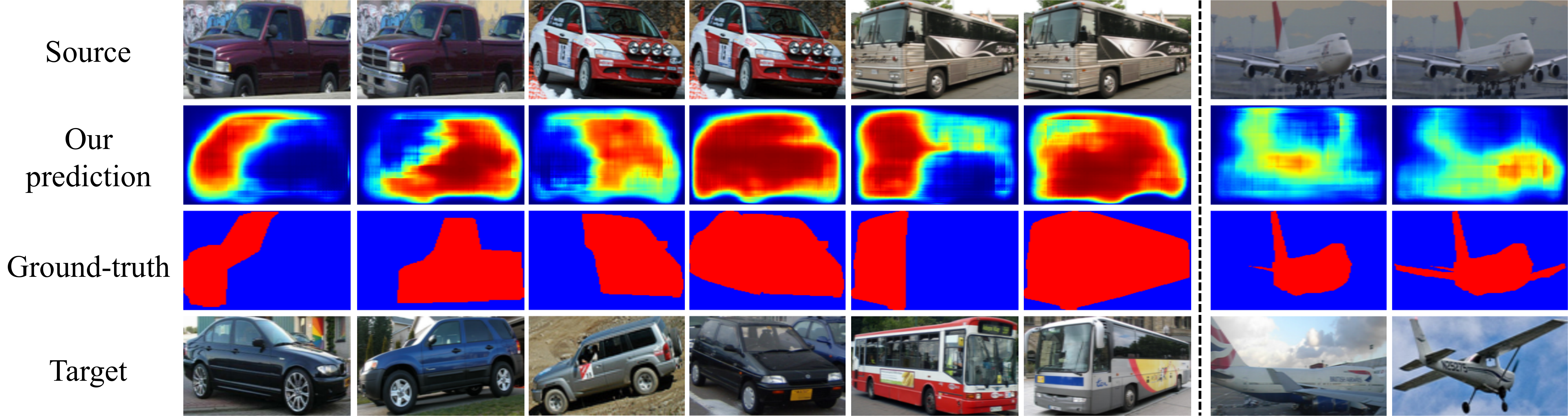}
\caption{Sample visualization of our matchability prediction. Notice how the prediction varies for the same source image when changing only the target image. The last two columns demonstrate a typical failure mode of our network having trouble localizing the fine boundaries of the matchable regions.}
\label{fig:match}
\end{figure*}

\subsection{Matchability prediction}
\lblsec{exp:match}
We evaluate the performance of matchability prediction using the PASCAL-Part dataset~\cite{chen2014detect}, which provides human-annotated part segment labeling\footnote{For categories without part labels, including boat, chair, table and sofa, we use the foreground segmentation mask instead.}. For each test image pair, a pixel in the source image is deemed matchable if there exists another pixel in the target image that shares the same part label, and all background pixels are unmatchable. We measure the performance by computing the percentage of pixels being classified correctly. For our method, we classify a pixel as matchable if its probability is $> 0.5$ according to the network prediction. To obtain matchability prediction for SIFT flow, we compute the $L_1$ norm of the SIFT feature matching error for each source pixel after the alignment, and a pixel is predicted to be matchable if the error is below a certain threshold (we did grid search on the training set to determine the threshold, and found $1,000$ to perform the best). Table~\ref{tab:match} compares the classification accuracy between our method and SIFT flow prediction (chance performance is $50\%$). Our method significantly outperforms SIFT flow on all categories except ``bicycle'' and ``motorbike'' ($67.8\%$ vs. $57.1\%$ mean accuracy).

We visualize some examples of our matchability prediction in Figure~\ref{fig:match}. Notice how the prediction varies when the target image changes with the source image being the same. 

\begin{table*}[t]
\centering
\scalebox{0.88}{
\begin{tabular}{ccccccccccccc|c}
\toprule
& aero & bike & boat & bottle & bus & car & chair & table & mbike & sofa & train & tv & {\bf mean} \tabularnewline
 \midrule
SIFT flow~\cite{liu2011sift}
& $66.2$ & $\mathbf{62.7}$ & $49.5$ & $50.5$ & $52.0$ & $64.5$ & $50.7$ & $50.5$ & $\mathbf{80.6}$ & $49.6$ & $58.5$ & $50.2$ & $57.1$ \tabularnewline
Ours
& $\mathbf{75.8}$ & $61.0$ & $\mathbf{66.7}$ & $\mathbf{67.1}$ & $\mathbf{67.3}$ & $\mathbf{72.0}$ & $\mathbf{66.1}$ & $\mathbf{68.4}$ & $68.0$ & $\mathbf{71.2}$ & $\mathbf{64.4}$ & $\mathbf{65.1}$ & $\mathbf{67.8}$ \tabularnewline
\bottomrule
\end{tabular}
}
\vspace{0.07in}
\caption{Performance comparison of matchability prediction between SIFT flow and our method (higher is better). See~\refsec{exp:match} for more details on the experiment setup.}
\label{tab:match}
\end{table*}

\subsection{Shape-to-image segmentation transfer}
\lblsec{app}
\begin{figure}[t]
\centering
\includegraphics[width = \columnwidth]{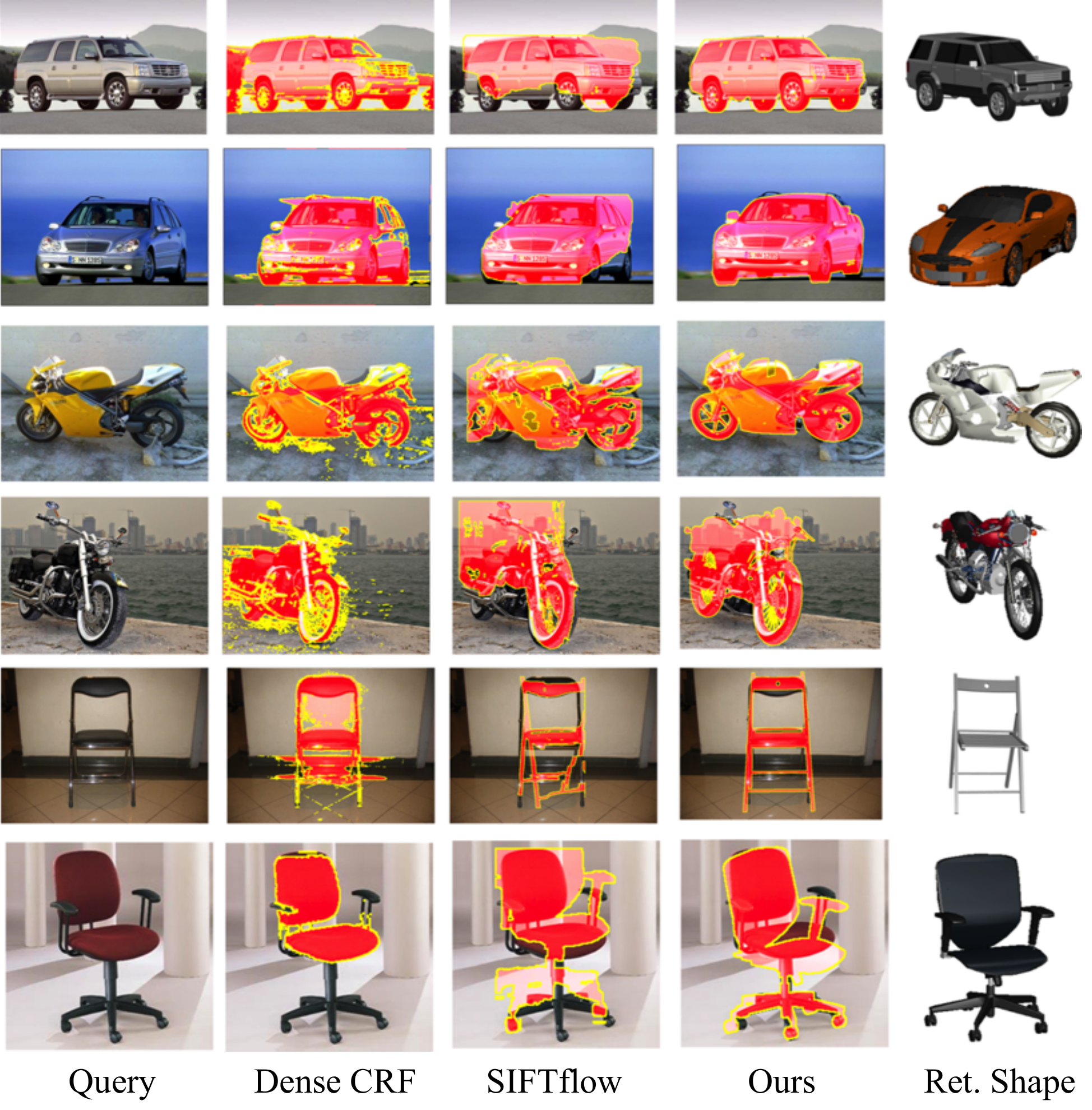}
\caption{Visual comparison among different segmentation methods. From left to right: input query image, segmentation by \cite{Krahenbuhl_Koltun_2011}, segmentation transferred using SIFT flow, segmentation transferred using our flow and the retrieved shape whose segmentation is used for transferring. See~\refsec{app} for more details.}
\label{fig:segmentation}
\end{figure}

Although in this paper we are mostly interested in finding correspondence between real images, a nice byproduct of our consistency training is that the network also implicitly learns cross-domain, shape-to-image correspondence, which allows us to transfer per-pixel labels (e.g. surface normals, segmentation masks, etc.) from shapes to real images. As a proof of concept, we ran a toy experiment on the task of segmentation transfer. Specifically, we construct a shape database of about $200$ shapes per category, with each shape being rendered in $8$ canonical viewpoints. Given a query real image, we apply our network to predict the correspondence between the query and each rendered view of the same category, and warp the query image according to the predicted flow field. Then we compare the HOG Euclidean distance between the warped query and the rendered views, and retrieve the rendered view with minimum error whose correspondence to the query image on the foreground region is used for segmentation transfer. Figure~\ref{fig:segmentation} shows sample segmentation using different methods. We can see that our learned flows tend to produce more accurate segmentation transfer than SIFT flow using the same pipeline. In some cases our output can even segment challenging parts such as the bars and wheels of the chairs.

\section{Discussion}
In this paper, we used cycle-consistency as a supervisory signal to learn dense cross-instance correspondences.  Not only did we find that this kind of supervision is surprisingly effective, but also that the idea of learning with cycle-consistency could potentially be fairly general. 
One could apply the same idea to construct other training scenarios, as long as the ground-truth of one or more edges along the cycle is known. We hope that this work will inspire more efforts to tackle tasks with little or no direct labels by exploiting cycle consistency or other types of indirect or ``meta''-supervision.

\section*{Acknowledgements}
We thank Leonidas Guibas, Shubham Tulsiani, and Saurabh Gupta for helpful discussions. This work was sponsored in part by NSF/Intel VEC 1539099, ONR MURI N000141010934, and a hardware donation by NVIDIA. 

{\small
\bibliographystyle{ieee}
\bibliography{egbib}
}

\end{document}